%% file: LinkPred.tex
\documentclass[conference]{IEEEtran}
 \pdfoutput=1
\RequirePackage{voila}
\RequirePackage{algorithmic}
\RequirePackage{epsfig}
\RequirePackage{graphicx}
\RequirePackage{changepage}
\usepackage{flushend}
\usepackage{multirow}

\usepackage{changepage}
\newlength{\graphicwidth}
\newlength{\smallgraphicwidth}
\begin{document}

\def\OurAlgo{CCLL}

\title{ \vspace*{-0.7cm}
Discriminative Link Prediction using Local Links, Node Features and Community Structure}
\author{Abir De \and Niloy Ganguly \and Soumen Chakrabarti}
\author{\IEEEauthorblockN{Abir De }
\IEEEauthorblockA{IIT Kharagpur, India\\
abir.de@cse.iitkgp.ernet.in
\vspace*{-0.5cm}}
\and
\IEEEauthorblockN{Niloy Ganguly}
\IEEEauthorblockA{IIT Kharagpur, India\\
niloy@cse.iitkgp.ernet.in \vspace*{-0.5cm}}
\and
\IEEEauthorblockN{Soumen Chakrabarti}
\IEEEauthorblockA{IIT Bombay, India\\
soumen@cse.iitb.ac.in}
\vspace*{-0.5cm}}
\maketitle
\input{Abs}
\input{Intro}
\input{Related}
\input{Eval}
\input{Discrim}

\input{Expt}

\input{End}
\vspace{0.3cm}
\input{Ack}
\bibliographystyle{abbrv}
\bibliography{voila}

\end{document}

%% file: Abs.tex
\begin{abstract}
A link prediction (LP) algorithm is given a graph, and has to rank, for each node, other nodes that are candidates for new linkage.  LP is strongly motivated by social search and recommendation applications.  LP techniques often focus on global properties (graph conductance, hitting or commute times, Katz score) or local properties (Adamic-Adar and many variations, or node feature vectors), but rarely combine these signals.  Furthermore, neither of these extremes exploit link densities at the intermediate level of communities.  In this paper we describe a discriminative LP algorithm that exploits two new signals.  First, a co-clustering algorithm provides community level link density estimates, which are used to qualify observed links with a surprise value.  Second, links in the immediate neighborhood of the link to be predicted are not interpreted at face value, but through a local model of node feature similarities.  These signals are combined into a discriminative link predictor.  We evaluate the new predictor using five diverse data sets that are standard in the literature.  We report on significant accuracy boosts compared to standard LP methods (including Adamic-Adar and random walk).  Apart from the new predictor, another contribution is a rigorous protocol for benchmarking and reporting LP algorithms, which reveals the regions of strengths and weaknesses of all the predictors studied here, and establishes the new proposal as the most robust.
\end{abstract}


%% file: Intro.tex
\section{Introduction}
\label{sec:Intro}

The link prediction (LP) problem \cite{LibenNowellK2007LinkPred} is to
predict future relationships from a given snapshot of a social
network.  E.g., one may wish to predict that a user will like a movie
or book, or that two researchers will coauthor a paper, a user will
endorse another on LinkedIn, or two users will become ``friends'' on
Facebook.  Apart from the obvious recommendation motive, LP can be
useful in social search, such as Facebook Graph
Search\footnote{\protect{\url{https://www.facebook.com/about/graphsearch}}},
as well as ranking goods or services based on not only real friends'
recommendations but also that of imputed social links.

Driven by these strong motivations, LP has been intensively researched
in recent years; Lu and Zhou \cite{LuZ2010LinkPred} provide a comprehensive
survey.  As we shall describe in Section~\ref{sec:Related},
in trying to predict if nodes $u,v$ in a social network are
likely to be(come) related, LP approaches predominantly exploit three
kinds of signals:
\begin{itemize}
\item When nodes have associated feature vectors (user demography,
  movie genre), node-to-node similarity may help predict linkage.
\item Local linkage information, such as the existence of many common
  neighbors, may hint at linkage.  The well-known Adamic-Adar (AA)
  \cite{AdamicA2003FriendsNeighbors} predictor and variants use such information.
\item Non-local effects of links, such as effective conductance,
  hitting time, commute time \cite{DoyleS1984RandomWalk}, or their
  heuristic approximations are often used as predictors.  The Katz
  score \cite{Katz1953Status} is a prominent example.  The random walk
  paradigm has also been combined \cite{BackstromL2011SRW} with edge
  features for enhanced accuracy.
\end{itemize}

Recently, stochastic block models \cite{HoEX2012TextLinks},
factor models and low-rank matrix factorization
\cite{LeeS2001NMF} have been used to ``explain'' a dyadic relation
using a frugal generative model.  These have rich connections to coding
and compression.  Co-clustering \cite{DhillonMM2003cocluster} and
cross-association \cite{ChakrabartiPMF2004CrossAssoc} are related
approaches.  Co-clustering exposes rich block structure in a
dyadic\footnote{Can be extended to larger arity.} relation.  E.g., in
a user-movie matrix, it can reveal that some users like a wide variety
of movies whereas others are more picky, or that some classic movies
are liked by all clusters of people.

Although co-clustering provides a \emph{regional} community density
signal, it is derived of global linkage considerations, and is
arguably more meaningful than global, unbounded random walks.
However, exploiting the signal from co-clustering is non-obvious.
The generative model implicit in co-clustering is that edges
in each dyadic block are sampled iid from a Bernoulli distribution
with a parameter corresponding to the empirical edge density in the
block.  While the choices of blocks and their edge densities offer
optimal global compression, they cannot predict presence or absence of
\emph{individual links} without incorporating node features and local
linkage information.

Our key contribution (Section~\ref{sec:Discrim}) is a new two-level
learning algorithm for LP.  At the lower level, we learn a local model
for similarity of nodes across edges (and non-edges).  This is
combined, using a support vector machine, 
with an entirely new non-local signal: the
output of co-clustering \cite{DhillonMM2003cocluster}, suitably
tuned into feature values.  

To the best of our knowledge, this is the
first work that brings all three sources of information (node
features, immediate neighborhood, ``regional'' community structure)
together in a principled way.  Another contribution
(Section~\ref{sec:Eval}) is a rigorous evaluation protocol for LP
algorithms, using a new binning strategy for nodes based on their
local link structure.  The new protocol reveals the strengths and
weaknesses of LP algorithms across a range of operating conditions.

On five diverse and public data sets (NetFlix, MovieLens, CiteSeer,
Cora, WebKb) that are standard in the LP community, our algorithm
offers (Section~\ref{sec:Expt}) substantial accuracy gains beyond
strong baselines.  Our main experimental observations are:
\begin{itemize}
\item The local similarity model on edges beats baselines in some
  regions of the problem space, but is not overall significantly
  better.
\item The additional signal from communities, found through
  co-clustering, is strong and helpful.
\item A global discriminative learning technique using the above signals
  convincingly beats baselines.
\end{itemize}


%% file: Related.tex
\section{Related work}
\label{sec:Related}

LP has been studied in different guises for many years, and
formalized, e.g., by Liben-Nowell and Kleinberg
\cite{LibenNowellK2007LinkPred}.  Lu and Zhou \cite{LuZ2010LinkPred}
have written a comprehensive survey.

\subsection{Local similarity}

If each node $u$ is associated with a feature vector $\theta_u$, these
can be used to define edge feature vectors $f(u,v) = f(\theta_u,
\theta_v)$, which can then be involved in an edge prediction through
logistic regression (i.e., $\Pr(\text{edge}|u,v) = \frac{1}{1 +
  e^{-\nu \cdot f(u,v)}}$), or a SVM (predict an edge if $\nu \cdot
f(u,v) > 0$).  Obviously, this class of models miss existing
neighborhood information.

To decide if nodes $u$ and $v$ may get linked, one strong signal is
the number of common neighbors they already share.  Adamic and Adar
(AA) \cite{AdamicA2003FriendsNeighbors} refined this by a weighted counting: common
neighbors who have many other neighbors are dialed down in importance:
\begin{align}
\similarity^{\text{AA}}_{i,j} = \sum_{k\in \Gamma(i)\cap \Gamma(j)}\frac{1}{\log d(k)},
\label{aa}
\end{align}
where $d(k)$ is the degree of common neighbor~$k$.

The resource allocation (RA) predictor is a slight variation which
replaces $\log d(k)$ with $d(k)$ in \eqref{aa}.  The AA and RA
predictors both penalize the contribution of high-degree common
neighbors.  The difference between them is insignificant if $d(k)$ is
small.  However they differ for large $d(k)$.  RA punishes the
high-degree common neighbors more heavily than AA.  Among ten local
similarity measures, Lu and Zhou found \cite[Table~1,
  page~9]{LuZ2010LinkPred} RA to be the most competitive, and AA
was close.

\subsection{Random walks and conductance}

AA, RA, etc.\ are specific examples of three general principles that
determine large graph proximity between nodes $u$ and $v$:
\begin{itemize}
\item Short path(s) from $u$ to $v$.
\item Many parallel paths from $u$ to $v$.
\item Few distractions (high degree nodes) on the path(s).
\end{itemize}
Elegant formal definitions of proximity, that capture all of the
above, can be defined by modeling the graph $(V,E)$ ($|V|=N, |E|=M$)
as a resistive network and measuring effective conductance, or
equivalently \cite{DoyleS1984RandomWalk}, modeling random walks
\cite{LangvilleM2004pagerank, TongFP2006restart, TongFK2007escape} on
the graph and measuring properties of the walk.  Many link predictors
are based on such proximity estimates.  The earliest, from 1953
\cite{Katz1953Status}, defines
\begin{align}
  \similarity_{\text{Katz}}(u,v) &= \beta A_{uv} + \beta^2 (A^2)_{uv} + \cdots
= (\mathbb{I} - \beta A)^{-1} - \mathbb{I},
\end{align}
where $A$ is the (symmetric) adjacency matrix and $\beta <
1/\lambda_1$, the reciprocal of the largest eigenvalue of~$A$.  This
is effectively a length-weighted count of the number of paths between
$u$ and $v$.  Lu and Zhou \cite{LuZ2010LinkPred} describe several
other related variants.  In their experiments, the best-performing
definition were local \cite{TongFP2006restart} and cumulative (called
``superposed'' by Lu and Zhou) random walks (LRW and CRW), described
next.

Suppose $q_u$ is the steady state visit
probability of node $u$ (degree divided by twice the number of edges
in case of undirected graphs).  Let $\pi_u(0)$ be the impulse
distribution at $u$, i.e., $\pi_u(0)[u] = 1$ and $\pi_u(0)[v] = 0$ for
$v\ne u$.  Define $\phi_u(t+1) = C^\top \pi_u(t)$, where $C$ is the $N
\times N$ row-stochastic edge conductance (or transition probability)
matrix.  Then
\begin{align}
\similarity_{uv}^{\text{LRW}}(t) &= q_u \pi_u(t)[v] + q_v \pi_v(t)[u].
\end{align}
For large $t$, the two rhs terms become equal, and LRW similarity is
simply twice the ``flow'' of random surfers on the edge $\{u,v\}$.  Lu
and Zhou claimed \cite[Table~3, page 16]{LuZ2010LinkPred} that LRW is
competitive, but the following cumulative random walk (CRW) is
sometimes more accurate.
\begin{align}
\similarity_{u,v}^{\text{CRW}}(t)
&=\sum_{\tau=1}^{t} \similarity_{u,v}^{\text{LRW}}(\tau). \label{srw}
\end{align}
CRW does not converge with increasing $t$, so $t$ is chosen by
validation against held-out data.

Although some of these approaches may feel ad-hoc, they work well in
practice; Sarkar \etal~\cite{SarkarCM2011LinkPred} have given
theoretical justification as to why this may be the case.

\subsection{Probabilistic generative models}

One of the two recent approaches that blend node features with linkage
information is by Ho \etal~\cite{HoEX2012TextLinks}, although it is
pitched not as a link predictor, but as an algorithm to cluster
hyperlinked documents into a Wikipedia-like hierarchy.  (Documents
directly correspond to social network nodes with local features.)  The
output is a tree of latent topic nodes, with each document associated
with a leaf topic.  The algorithm seeks to cluster similar documents
into the same or nearby topic nodes, and reward topic trees with dense
linkages between documents belonging to small topic subtrees rather
than span across far-away topic nodes.  The model associates a
parameter $\phi(t) \in (0,1)$ with each topic node $t$.  If documents
$u, v$ are attached to topic nodes $t(u), t(v)$, then the probability
of a link between $u,v$ is estimated as $\phi(\text{LCA}(t(u),t(v)))$,
where LCA is least common ancestor.  These probabilities can then be
used to rank proposed links.

\subsection{Supervised random walk (SRW)}

The other approach to blend node features with graph structure is
supervised and discriminative \cite{BackstromL2011SRW}, and based on
personalized PageRank \cite{JehW2003personalized}.  Recall that
$f(u,v)$ is an edge feature vector.  The raw edge weight is defined as
$a(u,v) = a(w \cdot f(u,v))$ for a suitable monotone function
$a(\cdot)>0$.  The probability of a random surfer walking from $u$ to
$v$ is set to
\begin{align*}
  \Pr(u \rightarrow v) \stackeq{\text{def}} C(u\rightarrow v)
  = C[v,u] = \frac{ a(u,v) } {\sum_{v'} a(u,v')}.
\end{align*}
where $C \in \R^{N\times N}$ is called the \emph{edge conductance
  matrix}.  If we are trying to predict out-neighbors of source node
$s$, we set up a teleport vector $r_s$ defined as $r_s[u] = 1$ if
$u=s$ and $0$ otherwise, then find the personalized PageRank vector
$\PPV_s$, defined by the recurrence
\begin{align*}
  \PPV_s = \alpha C \; \PPV_s + (1-\alpha) r_s.
\end{align*}
During training, for source node $s$, we are given some actual
neighbors $g$ and non-neighbors $b$, and we want to fit $w$ so that
$\PPV_s[g] > \PPV_s[b]$.

SRW is elegant in how it fits together edge features with visible
graph structure, but the latter is exploited in much the same way as
Katz or LRW.  Specifically, it does not receive as input regional
graph community information.  Thus, SRW and our proposal exploit
different sources of information.  Unifying SRW with our proposal is
left for future work.

\subsection{Unsupervised random walk}

Before SRW, Lichtenwalter \etal~\cite{LichtenwalterLC2010PropFlow}
introduced an unsupervised prediction method, PropFlow, which
corresponds to the probability that a restricted random walk starting
at $v_i$ ends at $v_j$ in some pre-specified $l$ steps or fewer using
link weights as transition probabilities.  The restrictions are that
the walk terminates upon reaching $v_j$ or upon revisiting any other
node.  This produces a score $s_{ij}$ which is used to predict new
links.  PropFlow is somewhat similar to rooted PageRank, but it is a
more localized measure of propagation, and is insensitive to
topological noise far from the source node.  The salient features are:
1.~the walk length is limited by $l$ which is very small (4--5), so
PropFlow is much faster than SRW, and 2.~unlike PageRank, it does not
need restarts and convergence but simply employs a modified breadth
search method.

\subsection{Co-clustering}

Similar documents share similar terms, and vice versa.  In general,
clustering one dimension of a dyadic relation (represented as a binary
matrix $A$, say) is synergistic with clustering the other.  Dhillion
\etal~\cite{DhillonMM2003cocluster} proposed the first
information-theoretic approach to group the rows and columns of $A$
into separate row and column \emph{groups} (also called \emph{blocks},
assuming rows and columns of $A$ have been suitably permuted to make
groups occupy contiguous rows and columns) so as to best compress $A$
using one link density parameter per (row group, column group) pair.

As we shall see, co-clustering and the group link densities can
provide information of tremendous value to link prediction algorithms,
of a form not available to AA, RA, LRW or CRW.  In recommending movies
to people, for instance, there are clusters of people that like most
movies, and there are clusters of classic movies that most people
like.  Then there are other richer variations in block densities.
Given a query edge in the LP setting, it seems natural to inform the
LP algorithm with the density of the block containing the query edge.

However, the estimated block density is the result of a global
optimization, and cannot directly predict one link.  That requires
combining the block density prior with local information (reviewed
above).  That is the subject of Section~\ref{sec:Discrim}.



%% file: Eval.tex
\section{Evaluation protocol}
\label{sec:Eval}

As described informally in Section~\ref{sec:Discrim:Definition}, a LP
algorithm applied to a graph snapshot is successful to the extent that
\emph{in future}, users accept high-ranking proposed links.  In
practice, this abstract view quickly gets murky, especially for graphs
without edge creation timestamps, but also for those with timestamps.
In this section we discuss the important issues guiding our evaluation
protocol and measurements.

\subsection{Labeling vs.\ ranking accuracy}
\label{sec:Eval:LabelVsRank}

Regarding the LP algorithm's output as a binary prediction (edge
present/absent) for each node pair, comparing with the true state of
the edge, and counting up the fraction of correct decisions, is a bad
idea, because of extreme skew in typical sparse social graphs: most
potential edges are absent.  The situation is similar to ranking in
information retrieval (IR) \cite{Liu2009LearningToRank}, where, for
each query, there are many fewer relevant documents than irrelevant
ones.  In LP, a separate ranking is produced for each node $q$ from a
set of nodes $Q$, which are therefore called \emph{query nodes}.

Fix a $q$ and consider the ranking of the other $N-1$ nodes.  Some of
these are indeed neighbors (or will end up becoming neighbors).
Henceforth, we will call $q$'s neighbors as \emph{good} nodes $G(q)$
and non-neighbors as \emph{bad} nodes $B(q)$.  Ideally, each good node
should rank ahead of all bad nodes.  Because the LP algorithm is
generally imperfect, there will be exceptions.  The area under the ROC
curve (AUC) is widely used in data mining as a accuracy measure
somewhat immune to class imbalance.  It is closely related to the
fraction of the $|G(q)|\,|B(q)|$ good-bad pairs that are in the
correct order in LP's ranking.  However, for the same reasons as in IR
ranking \cite{Liu2009LearningToRank}, AUC tends to be large and
undiscerning for almost any reasonable LP algorithm.  Therefore, we
adapt standard ranking measure mean average precision (MAP).

At each node, given a score from the LP algorithm on all other nodes
as potential neighbors, and the ``secret'' knowledge of who is or
isn't a neighbor, we compute the following performance metrics.

\subsubsection{Precision and recall}

These are defined as
\begin{align}
Precision(k) &=\frac{1}{|Q|}\sum_{q \in Q} P_{q}(k) \label{eq:MeanPrec} \\
\text{and} \quad Recall(k) &= \frac{1}{|Q|}\Sigma_{q\in Q} R_q(k),
\label{eq:MeanRecall}
\end{align}
where $|Q|$ is the number of queries, $P_{q}(k)$ is Precision@$k$ for
query $q$, and $R_{q}(k)$ is Recall@$k$ for query~$q$.  So
$Precision(k)$ is the average of all Precision@$k$ values over the
set of queries, and likewise with $Recall(k)$.

\subsubsection{Mean average precision (MAP)}

First we define at query node $q$ the quantity
\begin{align}
AvP(q)=\frac{1}{L}\sum_{k=1}^{N-1}P_{q}(k) \; r_{q}(k)
\label{eq:MAP}
\end{align}
at each node, where $N-1$ is the number of nodes excluding the query
node itself, $L$ is the number of retrieved relevant items and
$r_{i}(k)$ is an indicator taking value 1 if the item at rank $k$ is a
relevant item (actual neighbor) or zero otherwise (non-neighbor).
Averaging further over query nodes, we obtain
$MAP=\frac{1}{|Q|}\sum_{q} AvP(q)$.

In the remainder of this section, we explore these important issues:
\begin{itemize}
\item How should $Q$ be sampled from $V$?  A related question is, how
  to present precision, recall, MAP, etc., with $Q$ suitably
  disaggregated to understand how different LP algorithms behave on
  different nodes $q\in Q$?  (See Section~\ref{sec:Eval:QuerySample}.)
\item Many graphs do not have link timestamps.  For these, once $Q$ is
  set, how should we sample edges incident on $Q$ for training and
  testing?  (See Section~\ref{sec:Eval:EdgeSample}.)
\end{itemize}

\subsection{Query node sampling protocols}
\label{sec:Eval:QuerySample}

In principle, the ranking prowess of an LP algorithm should be
evaluated at \emph{every} node.  E.g., Facebook may recommend friends
to all users, and there is potential satisfaction and payoff from
every user.  In practice, such exhaustive evaluation is intractable.
Therefore, nodes are sampled; usually $|Q| \ll N$.  On what basis
should query nodes be sampled?  In the absence of the social network
counterpart to a commercial search engine's query log, there is no
single or simple answer to this question.  LP algorithms often target
predicting links that close triangles if/when they appear.  92\% of
all edges created on Facebook Iceland close a path of length two,
i.e., a triangle \cite{BackstromL2011SRW}. These nodes are sampled as query nodes $Q$. 

Besides providing comparison of overall performance averaged over
query nodes, in order to gain insight into the dynamics of different
LP algorithms, we need to probe deeper in the structure of the network
and check the strength/weakness of the algorithm vis-a-vis various
structures.  In our work, we bucket the selected query nodes based on
\begin{itemize}
\item the number of neighbors,
\item the number of triangles on which they are incident.
\end{itemize}

\subsection{Edge sampling protocol}
\label{sec:Eval:EdgeSample}

If edges in the input graph have creation timestamps, we can present a
snapshot to the LP algorithm and simulate further passage of time to
accrue its rewards.  Even this happier situation raises many
troublesome questions, such as when the snapshot it taken, the horizon
for collecting rewards, etc., apart from (the composition of) the
query node sample.

To complicate matters further, many popular data sets (some used in
Section~\ref{sec:Expt}) do not have edge timestamps.  One extreme way
to dodge this problem is the leave-one-out protocol: remove exactly
one edge at a time, train the LP algorithm, and make it score that
edge.  But this is prohibitively expensive.  

Rather than directly sample edges, we first sample query nodes $Q$ as mentioned in
Section~\ref{sec:Eval:QuerySample}.  This narrows our attention to $|Q|(N-1)$ potential
edge slots incident on query nodes.  Fix query $q$. In the
fully-disclosed graph, $V\setminus q$ is partitioned into ``good''
neighbors $G(q)$ and ``bad'' non-neighbors $B(q)$.  We set a train
sampling fraction $\sigma\in(0,1)$.  We sample $\lceil \sigma |G(q)|
\rceil$ good and $\lceil \sigma |B(q)| \rceil$ bad nodes and present
the resulting \emph{training} graph to the LP algorithm.  ($\sigma$ is
typically 0.8 to 0.9, to avoid modifying the density and connectivity
of the graph drastically and misleading LP algorithms.)

The good and bad training samples are now used to build our models as
described in Section~\ref{sec:Discrim}.  The training graph, with the
testing good neighbors removed, is used for
co-clustering.  This prevents information leakage from the training
set.  The remaining good neighbors and bad non-neighbors are used for
testing.  In case $|G(q)| = \lceil \sigma |G(q)| \rceil$ or $|B(q)| =
\lceil \sigma |B(q)| \rceil$, we discard $q$, introducing a small bias
after our sampling of $Q$.  Effectively this is a ``sufficient
degree'' bias, which is also found in prior art \cite[Section 4: $K,
  \Delta$]{BackstromL2011SRW}.


%% file: Discrim.tex
\section{Proposed framework: \OurAlgo}
\label{sec:Discrim}

We have reviewed in Section~\ref{sec:Related} several LP approaches.
Some (AA, RA, CRW) involve no learning, others \cite{HoEX2012TextLinks}
propose generative probabilistic models that best ``explain'' the
current graph snapshot (and then the model parameters of the
``explanation'' can be used to predict future links, although they did
not study this application).  In recent years, direct prediction of
hidden variables through conditional probability
\cite{LaffertyMP2001crf} or discriminative
\cite{TsochantaridisJHA2005structsvm} models have proved generally
superior to modeling the joint distribution of observed and hidden
variables~\cite{Vapnik1998learning}.  As we shall see in
Section~\ref{sec:Expt}, this is confirmed even among our comparisons
of prior work, where supervised random walk \cite{BackstromL2011SRW}
is superior to unsupervised approaches. However before explaining our
scheme we clearly define the \emph{link prediction} problem.

\subsection{Problem definition}
\label{sec:Discrim:Definition}

We are given a snapshot of a social network, represented as a graph
$(V,E)$ with $|V|=N$ and $|E|=M$.  Let $u \in V$ be a node (say
representing a person).  Edges may represent ``friendship'', as in
Facebook.  Depending on the application or algorithm, the graph may be
directed or undirected.  The goal of LP is to recommend new friends to
$u$, specifically, to rank all other nodes $v \in V\setminus u$ in
order of likely friendship preference for~$u$.  One ranking device is
to associate a \emph{score} with each $v$, and sort them by decreasing
score.  LP algorithms vary in how they assign this score.  We can also
think about LP as associating a binary hidden variable with the
potential edge $(u,v)$, then estimating the probability that this
variable has value 1 (or true), based on observations about the
currently revealed graph.  The LP algorithm is considered high quality
if the user accepts many proposed friends near the top of the ranked
list.  In case of large $V$, LP systems often restrict the potential
set of $v$s to ones that have a common neighbor $c$, i.e., $(u,c)$ and
$(c,v)$ already exist in~$E$.

\subsection{Overview of two-level discriminative framework}
\label{sec:Discrim:Overview}

LP can also be regarded as a classification problem: given a pair of
nodes $u,v$, we have two class labels (``link'' vs.\ ``no link'') and
the task is to predict the correct label.  To estimate a confidence in
the (non) existence of a link, we will aggregate several kinds of
input signals, described throughout the rest of this section.  Apart
from subsuming existing signals from one or more of AA, RA, and CRW,
we will harness two new signals.  First, in
Section~\ref{sec:Discrim:LocalLearning}, we will describe a local
learning process to determine effective similarity between two given
nodes.  Unlike AA and other node-pair signals, our new approach
recognizes that propensity of linkage is not purely a function of node
similarity; it changes with neighborhood.  Second, in
Section~\ref{sec:Discrim:CoclusterSurprise} we describe how to harness
the output of a co-clustering of the graph's adjacency matrix to
derive yet more features.  To our knowledge coclustering has never
been used in this manner for LP.

For each proposed node pair $u,v$, these signals will be packed as
features into a feature vector $f(u,v) \in \R^d$ for some suitable
number of features $d$.  We estimate a \emph{global model} $\nu \in
\R^d$, such that the confidence in the existence of edge $(u,v)$ is
directly related to $\nu \cdot f(u,v)$.  $f(u,v)$ will consist of
several blocks or sub-vectors, each with one or more elements.
\begin{itemize}
\item $f(u,v)[\text{AA}]$ is the block derived from the Adamic-Adar
  (AA) score \eqref{aa}.  (We can also include other local scores in
  this block.)
\item $f(u,v)[\text{LL}]$ is the block derived from local similarity
  learning (Section~\ref{sec:Discrim:LocalLearning}).
\item $f(u,v)[\text{CC}]$ is the block derived from co-clustering
  (Section~\ref{sec:Discrim:CoclusterSurprise}).
\end{itemize}
As we shall demonstrate in Section~\ref{sec:Expt}, these signals
exploit different and complementary properties of the network.  If
$y(u,v) \in \{0, 1\}$ is the observed status of a training edge, we
can find the weights ($\nu$) using a SVM and its possible variations.
The details will be discussed in Section~\ref{sec:Discrim:LCG}.

To exploit possible interaction between features, we can construct
suitable kernels \cite{Bishop2006PRML}.  Given we have very few
features, a quadratic (polynomial) kernel can be implemented by
explicitly including all quadratic terms.  I.e., we construct a new
feature vector whose elements are
\begin{itemize}
\item $f(u,v)[i]$ for all indices $i$, and
\item $f(u,v)[i]\, f(u,v)[j]$ (ordinary scalar product)
for all index pairs $i,j$.
\end{itemize}
We can also choose an arbitrary informative subset of these.
We will now describe the three blocks of features.

\subsection{Learning local similarity}
\label{sec:Discrim:LocalLearning}

An \emph{absolute} notion of similarity
between $u$ and $v$, based on node features $\theta_u, \theta_v$, is
not strongly predictive of linkage; it also depends on the typical
\emph{distribution} of similarity values in the
neighborhood~\cite{Doubleblind}.  
Also, the presence or absence of edge $(u,v)$ rarely determined by
nodes far from $u$ and $v$.  Keeping these in mind, the first
step of the algorithm learns the typical (dis)similarity between $u$
and $v$ and their common neighbors.  We term this the \emph{reference}
dissimilarity.  We then use this to predict the chance of link
$(u,v)$ arriving.

Let $\Gamma(u)$ be the (immediate) neighbors of $u$.
We will model the \emph{edge dissimilarity} between $u$ and $v$ as
\begin{align}
  \Delta_{w}(u,v) &= w_{uv} \cdot |\theta_{u}-\theta_{v}|,
\label{eq:Dissim}
\end{align}
where $\theta_u$ is a node feature vector associated with $u$, and
$|\cdots|$ denotes the \emph{elementwise} absolute value of a vector,
e.g., $|(-2, 3)|=(2, 3)$, although other general combinations of
$\theta_u$ and $\theta_v$ are also possible \cite{BackstromL2011SRW}.
$w_{uv}$ is the weight vector fitted \emph{locally} for $u,v$.
(Contrast this with the global $\nu$ above, and the final proposal in
SRW \cite{BackstromL2011SRW} that fits a single model over all node
pairs.)

\subsubsection{Finding $w_{uv}$ and reference dissimilarity}

Throughout this work, and consistent with much LP work, we assume that
edges are associative or unipolar, i.e., there are no ``dislike'' or
antagonistic links.  Similar to AA and friends, when discussing node
pair $u,v$, we restrict our discussion to the vicinity
$N=\Gamma(u)\cup \Gamma(v)$.

For $A \subseteq V \setminus u$, $A \ne \varnothing$, we extend
definition \eqref{eq:Dissim} to the \emph{set dissimilarity}
\begin{align}
 \Delta_w(u,A) &= \frac{1}{|A|} \sum_{v \in A} \Delta_w(u,v).
\label{eq:SetDissim}
\end{align}
We define $\Delta_w(u,\varnothing) = 0$.
$\Delta_w(u,A)$ is the average dissimilarity between $u$ and $A$.
Note that $\Delta_w(u, \{v\})$ is simply $\Delta_w(u,v)$.

\begin{figure}[th]
    \centering
    \includegraphics[width=\hsize]{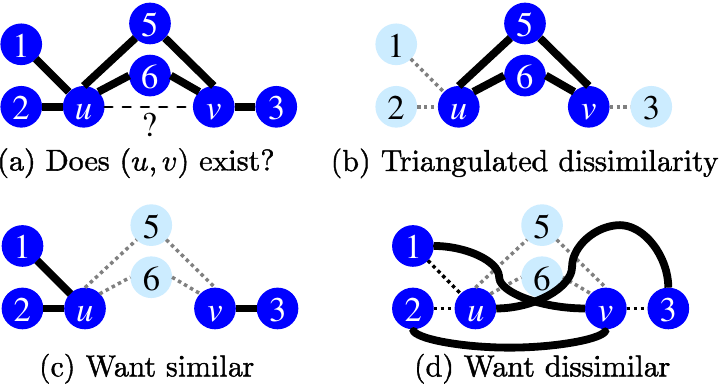}
\caption{Local dissimilarity constraints.}
    \label{fig:CllExample}
\end{figure}

The key idea here is that, if there is an edge $(u,v)$, we want to
choose $w_{u,v}$ such that $\Delta_w(u,v)$ is low, \emph{relative to}
node pairs that are not neighbors.  Conversely, if $(u,v)$ is not an
edge, we want the dissimilarity to be large relative to nearby node
pairs that are neighbors.  We codify this through the following four
constraints:
\begin{align}
\Delta_w(u, \Gamma(u) \setminus \Gamma(v)) &\le \alpha \Delta_w(u,v) \notag \\
\Delta_w(v, \Gamma(v) \setminus \Gamma(u)) &\le \alpha \Delta_w(u,v) 
\label{eq:CllConstraints} \\
\Delta_w(u, \Gamma(v) \setminus \Gamma(u)) &\ge \beta \Delta_w(u,v) \notag \\
\Delta_w(v, \Gamma(u) \setminus \Gamma(v)) &\ge \beta  \Delta_w(u,v) \notag
\end{align}
Figure~\ref{fig:CllExample} illustrates the constraints.
Here $\alpha, \beta$ are suitable multiplicative margin parameters.
Smaller (larger) value of $\alpha$ ($\beta$) allows a
lower (higher) dissimilarity between the connected (disconnected)
nodes.  Here, we have experimentally selected $\alpha$ and $\beta$.

Subject to the above constraints \eqref{eq:CllConstraints}, we wish to
choose $w$ so as to minimize $\Delta_w(u,v)$.  This is a standard
linear program, which, given the typically modest size of $\Gamma(u)
\cup \Gamma(v)$, runs quite fast.

\subsubsection{Computation of LL features}

The linear program outputs $w^*_{uv}$, from which we can compute
\begin{align}
  \delta_{uv} = w^*_{uv} \cdot |\theta_u - \theta_v|.
\end{align}
But is $\delta_{uv}$ larger than ``expected'', or smaller?  Plugging
in the raw value of $\delta_{uv}$ into our outer classifier may make
it difficult to learn a consistent model $\nu$ globally across the
graph.  Therefore, we also compute the \emph{triangulated
  dissimilarity} between $u$ and $v$, using common neighbors $i$, as
\begin{align}
  \bar\Delta_{w^*}(u,v) &= 
  \sum_{i \in \Gamma(u) \cap \Gamma(v)} \frac{\Delta_{w^*}(i,u) + \Delta_{w^*}(i,v)}
{|\Gamma(u) \cap \Gamma(v)|} \\
&= \Delta_{w^*}(u, \Gamma(u) \cap \Gamma(v)) +
\Delta_{w^*}(v, \Gamma(u) \cap \Gamma(v)).  \notag
\end{align}
Finally, we return
\begin{align}
  f(u,v)[LL] &= \bar\Delta_{w^*}(u,v) - \delta_{uv}.
\label{eq:LLfeature}
\end{align}
If $f(u,v)[LL]$ is large and positive, it expresses confidence that
link $(u,v)$ will appear; if it is large and negative, it expresses
confidence that it will not.  Around zero, the LL feature is
non-committal; other features may then come to the rescue.

\subsection{Co-clustering and ``surprise''}
\label{sec:Discrim:CoclusterSurprise}

Given a dyadic relation represented as a matrix, co-clus\-tering
\cite{DhillonMM2003cocluster} partitions rows and columns into row
groups and column groups.  We can think of the rows and columns of the
input matrix being reordered so that groups are contiguous.  The
intersection of a row group and column group is called a
\emph{block}. The goal of co-clustering is to choose groups so that
the blocks are \emph{homogeneous}, i.e., edges within a block appear
uniformly dense, with no finer structure (stripes) based on rows or
columns.  Co-clustering uses a data compression formalism to determine
the optimal grouping.

Consider a query node pair $u,v$ where we are trying to predict
whether edge $(u,v)$ exists.  E.g., $u$ may be a person, $v$ may be a
movie, and we want to predict if $u$ will enjoy watching $v$.  In this
person-likes-movie matrix, as a specific example, there may be row
groups representing clusters of people that like most movies, and
there may be column groups representing clusters of classic movies
that most people like.  In general, the block in which the matrix
element $[u,v]$ is embedded, and in particular, its edge density
$d(u,v)$, gives a strong prior belief about the existence (or
otherwise) of edge $(u,v)$, and could be the feature $f(u,v)[CC]$ in
and of itself.

Although block density $d(u,v) \in [0,1]$, the penalty for deviating
from it in the ultimate link decision is not symmetric (thanks again
to graph sparsity).  So a better 
formalism to capture a coclustering-based feature is the ``surprise
value'' of an edge decision for node pair $u,v$.  As an extreme case,
if a non-edge (value 0) is present in a co-cluster block where all
remaining elements are 1 (edges), it causes large surprise.  The same
is the case in the opposite direction.

There are various ways of expressing this quantitatively.  One way of
expressing it is that if an edge $(u,v)$ is claimed to exist, and
belongs to a block with an edge density $d(u,v)$, the surprise is
inversely related to $d(u,v)$; in information theoretic terms, the
surprise is $-\log d(u,v)$ bits.  (So if $d(u,v) \rightarrow 0$, yet
the edge exists, the surprise goes to $+\infty$.)  Similarly, if the
edge does not exist, the surprise is to $-\log(1-d(u,v))$ bits.


\subsection{The discriminative learner for global model $\nu$}
\label{sec:Discrim:LCG}

In order to obtain the best LP accuracy, the above signals need to be
combined suitably.  For each edge, there are two classes
(present/absent).  One possibility is to label these $+1, -1$, and fit
the predictor $\hat y_{uv} = \sign(\nu \cdot f(u,v))$.

\subsubsection{Loss function}
$\nu$ can be learnt to minimize various loss functions.  The simplest
is 0/1 edge misclassification\footnote{$\llbracket B \rrbracket$ is 1
  if $B$ is true, 0 otherwise.} loss $\sum_{q \in Q} \frac{1}{N-1}
\sum_{v \ne q} \llbracket \hat y_{qv} \ne y_{qv} \rrbracket$, which is
usually replaced by a convex upper bound, the hinge loss
\begin{align}
\sum_{q\in Q} \frac{1}{|G(q)\cup B(q)|} 
\hspace{-1em}
\sum_{v \in G(q) \cup B(q)}
\hspace{-1.5em}
\max\left\{ 0, y_{qy} \nu \cdot f(u,v) - 1 \right\}.
\end{align}
As we have discussed in Section~\ref{sec:Eval:LabelVsRank}, for
ranking losses, it is better to optimize the AUC, which is closely
related \cite{Joachims2005multivariate} to the pairwise loss
\begin{align}
  \sum_{q \in Q} \frac{1}{|G(q)|\,|B(q)|}
  \sum_{g \in G(q), b \in B(q)} \hspace{-1.5em} \llbracket
  \nu \cdot f(q,b) \ge \nu \cdot f(q,g) \rrbracket,
\end{align}
which is again usually approximated by the hinge loss
\begin{align}
  \sum_{q \in Q} \tfrac{1}{|G(q)|\,|B(q)|}
\hspace{-2em}
  \sum_{g \in G(q), b \in B(q)} \hspace{-2em} 
\max\bigl\{ 0, 1 - \nu \cdot (f(q,g)\! -\! f(q,b)) \bigr\}
\end{align}
Joachims \cite{Joachims2005multivariate} offers to directly optimize
$\Lambda$ for several ranking objectives; we choose area under the ROC
curve (AUC) for training $\Lambda$, although we evaluate the resulting
predictions using MAP.  During inference, given $q$, we find $\nu
\cdot f(q,v)$ for all $v\ne q$ and sort them by decreasing score.

\subsubsection{Feature map}

We now finish up the design of $f(u,v)[AA], f(u,v)[LL]$ and
$f(u,v)[CC]$.  The first two are straight-forward, we simply use the
single scalar \eqref{aa} for $f(u,v)[AA]$, and $f(u,v)[LL]$ is also a
single scalar as defined in \eqref{eq:LLfeature}.  The $f(u,v)[CC]$
case is slightly more involved, and has two scalar elements, one for
each surprise value:
\begin{itemize}
\item $-\log d(u,v)$ for the ``link exists'' case, and
\item $-\log (1 - d(u,v))$ for the ``edge does not exist'' case.
\end{itemize}
Accordingly, $\nu$ will have two model weights for the CC block, and
these will be used to balance the surprise values from training data.
The soundness of the above scheme follows from structured learning
feature map conventions
\cite{Joachims2005multivariate,TsochantaridisJHA2005structsvm}.  We
defer the elementary algebraic details to the full version of this
paper.  Thus, $f(u,v)$ has a total of four elements.


%% file: Expt.tex
\section{Experiments}
\label{sec:Expt}

We compare \OurAlgo\ against several strong baselines such as
Adamic-Adar (AA) \cite{AdamicA2003FriendsNeighbors}, RA
\cite{LuZ2010LinkPred}, Cumulative Random Walk (CRW)
\cite{LuZ2010LinkPred}, Supervised Random Walk (SRW)
\cite{BackstromL2011SRW}, and Generative Model (GM)
\cite{HoEX2012TextLinks}.  RA computes the score in a similar manner
like AA, and therefore gives almost same performance.  Therefore, we
omit RA and only present those for AA.  Apart from using LL as
features to \OurAlgo, we run LL independently as a baseline.

\subsection{{Datasets used}}

We used the following popular public data sets, also summarized in
Figure~\ref{tab:datasets}.
\begin{LaTeXdescription}
\item[Movielens \cite{ml}:] It has 6040 users and 3952 movies. Each
  user has rated at least one movie.  Each movie has features which
  are a subset of a set of 18 nominal attributes (e.g. animation,
  drama etc.).  From the raw data we constructed a ``social network''
  between movies where two movies have an edge if they have at least a
  certain number of common viewers.  By choosing the minimum number of
  common viewers to be 100, we obtain a network with 3952 nodes and
  5669 edges.
\item[CiteSeer \cite{ud}:] The CiteSeer dataset consists of 3312
  scientific publications and the citation network consists of 4732
  links. Each publication is tagged with a set of keywords. Total
  number of keywords is 3703.
\item[Cora \cite{ud}:] The Cora dataset consists of 2708 scientific
  publications and the citation network consists of 5429 links. Here
  the total number of keywords is 1433.
\item[WebKb \cite{ud}:] The WebKb dataset consists of 877 scientific
  publications and the citation network consists of 1608 links. Here
  the total number of keywords is 1703.
\item[Netflix \cite{netflixprize}:] The Netflix dataset consists of
  2649429 users and 17770 movies. Each user has rated at least one
  movie. Each movie has 64 features obtained by SVD from many factors.
  From the raw daya we constructed a ``social network'' of movies
  where two movies have an edge if they have at least a certain number
  of common viewers. By choosing the minimum number of common viewers
  to be 20, we obtain a network with 17770 nodes and 20466 edges.
\end{LaTeXdescription}

\begin{figure}
\renewcommand{\arraystretch}{1.3}
\centering \begin{tabular}{|c||c|c|c|c|}
\hline Dataset   &  N     &  E     & $n(a)$   &   $d_{avg}$ \\ \hline\hline Movielens &  3952  &  5669  &  18 &
2.8689      \\ \hline CiteSeer  &  3312  &  4732  &  3703   &         2.7391    \\ \hline Cora    &  2708  &
5429  & 1433    &      3.89  \\ \hline WebKb     &  877  &  1608   &  1703     &        2.45       \\ \hline
NetFlix   & 17770   &  20466   &  64   &         2.3034     \\ \hline
\end{tabular}
\caption{Summary of the datasets, where $N$ is the number of items,
  $E$ is the total number of links, $n(a)$ is the number of features
  and $d_{avg}$ is the average degree. }
\label{tab:datasets}
\end{figure}

\subsection{Performance of \OurAlgo\ compared to other algorithms}

\begin{figure*}
\centering \includegraphics[width=\textwidth]{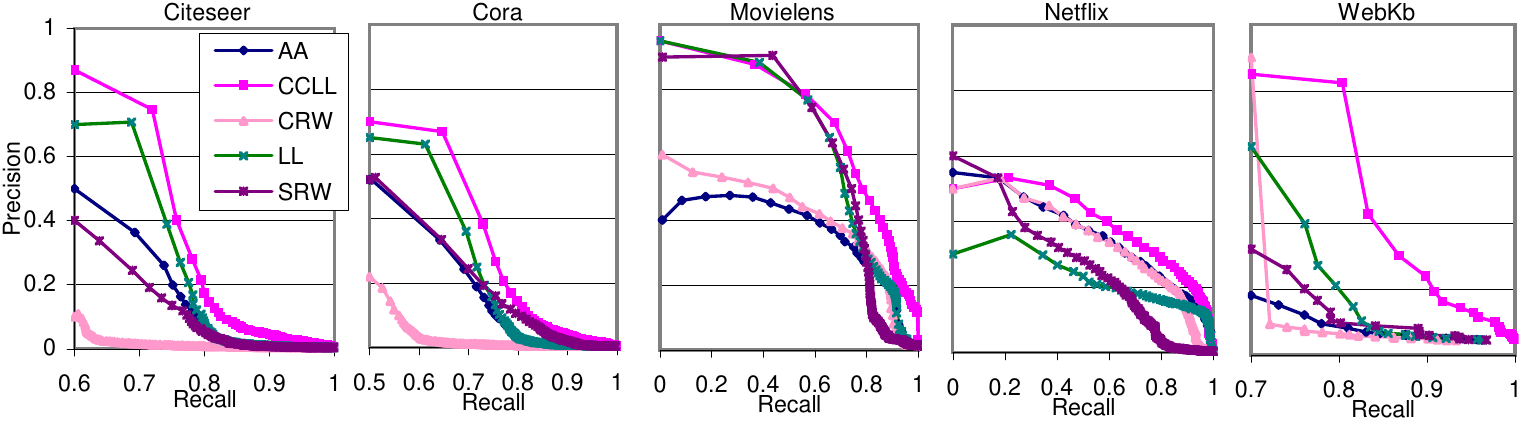}
    \caption{Precision vs.\ recall curves for all data sets and algorithms.}
    \label{fig:precision_recall}
\end{figure*}

Figure~\ref{tab:results} gives a comparative analysis of MAP (Mean
Average Precision) values for all datasets and algorithms, and
Figure~\ref{fig:precision_recall} gives a more detailed view of
precision vs.\ recall.  We observe that, for all datasets, the overall
performance of \OurAlgo\ is substantially better than all other
methods.

Performance of the probabilistic generative model is particularly
poor.  This was surprising to us, given stochastic block models (SBMs)
seem ideally suited for use in LP.  Closer scrutiny showed model
sparsity as a likely culprit, at least in case of Ho \etal's
formulation.  They derive a tree-structured hierarchical clustering of
the social network nodes, where the number of hierarchy nodes is much
smaller than $N$, the number of social network nodes.  Their model
assigns a score to an edge $(u,v)$ that depends on the hierarchy paths
to which $u$ and $v$ are assigned.  Since the number of hierarchy
nodes is usually much smaller than the number of social nodes, the
score of neighbors of any nodes have a lot of ties, which reduces
ranking resolution.  Therefore, MAP suffers.  In contrast, the coarse
information from co-clustering (CC) is only a feature into our
top-level ranker.

AA, RA, CRW all produce comparable performance.  All these methods
solely depend on link characteristics, for example AA and RA depending
on the number of triangles a node is part of, hence they miss out the
important node or edge feature information.  Regarding CRW, as $t$
goes to $\infty$, it doesn't converge, and there is no consistent
global $t$ for best MAP.  SRW, which performs best among the
baselines, uses node and link features (PageRank) but not community
based (co-clustering) signal.  Moreover, SRW learns only one global
weight vector, unlike $w_{uv}$ in LL, a signal readily picked up by
\OurAlgo.  We also found the inherent non-convexity of SRW to produce
suboptimal optimizations.  LL is next to SRW in all data sets except
NetFlix.  This is because the number of features in Netflix is small
and the values assumed by each feature is very diverse, making the
feature-based local prediction strategy ineffective.

A possible explanation of the superior performance of \OurAlgo\ can be
that the underlying predictive modules (LL, AA, CC) perform well in
complementary zones which \OurAlgo\ can aggregate effectively.  In
order to probe into this aspect we make a detailed study of the
performance of the various algorithms with respect to various
distribution of work load (elaborated in Section~\ref{workLoad}).

\if{0} In fact in our initial poster paper \cite{abirrecsys} there has been an indication that AA and LL
perform well in complementary zones. Particularly, where the nodes with high degree are connected together,
Adamic-Adar metric fails to provide high score. So a probable reason is that  we are been able to combine this
seemingly complementary schemes. \fi

\begin{figure}
\begin{adjustwidth}{-0.5cm}{}
\begin{tabular}{|p{1.3cm}||c|c|c|c|c|c|}
\hline Dataset   & \OurAlgo\  &  LL  & AA &CRW & GM &SRW\\ \hline\hline
 Netflix          & 0.6017 &  0.4750  &  0.5381  &  0.5428 &  0.1268&0.4564   \\ \hline
Movielens    & 0.8747 &  0.8133  &  0.5341 &  0.5784  &  0.20  &0.8114 \\ \hline 
CiteSeer         & 0.7719 &  0.7393  & 0.6649  &  0.5309  &  0.1452    &0.6281 \\ \hline
Cora             & 0.7234 & 0.6805  &0.6135 &    0.4726&0.0583&0.6274 \\ \hline
WebKb     & 0.8583 &  0.7505 &  0.6035   &  0.5736  &  0.3360  &0.6693 \\ \hline
\end{tabular}
\caption{Mean average precision over all algorithms and datasets.}
\label{tab:results}
\end{adjustwidth}
\end{figure}


\subsection{Stability to Sampling}

Among all methods, \OurAlgo\ and SRW use machine learning. Hence in
order to train the model, we randomly select a certain fraction of
edges (say $TS\%$) and the same fraction of non-edges from the
network.  CRW is an unsupervised algorithm but since it performs a
global random walk it is affected by the sampling, the performance
deteriorating when many edges are removed.  Figure~\ref{tab:sampling}
shows the variation of performance with training sets of different
sizes.  We conducted the experiment with 80\% and 90\% training
samples.  When we decrease the sample size from 90\% to 80\%, the
performance deteriorates for all methods.  But the deterioration is
much smaller in \OurAlgo\ compared to CRW and SRW.  This is because
\OurAlgo\ picks up strong signals from the AA and LL features.  AA and
LL work solely on local factors.  Deletion of an additional 10\% of
edges hardly affects AA or LL score.

\begin{figure}
\centering\begin{tabular}{|p{1.4cm}|c|c|c|c|}
\hline%
 Dataset &TS (\%)& \OurAlgo\  &CRW&SRW \\ \hline
\multirow{2}{12mm}{\centering Netflix}&80&0.5263  &0.4493 & 0.3849 \tabularnewline
 \cline{2-5}%
 & 90& 0.6017  &0.5428 & 0.5567 \tabularnewline
\hline%
\multirow{2}{12mm}{\centering Movielens}&80 &0.8672  &0.5018 & 0.7789 \tabularnewline
 \cline{2-5}%
&90 & 0.8740   &0.5784 & 0.8114 \tabularnewline
\hline%

\multirow{2}{12mm}{\centering Citeseer}&80&0.7000 &0.3295 & 0.5567 \tabularnewline
 \cline{2-5}%
 & 90& 0.7719 &0.5309 & 0.6281 \tabularnewline
\hline%

\multirow{2}{12mm}{\centering Cora}&80&0.6803  &0.3400& 0.5516  \tabularnewline
 \cline{2-5}%
 & 90& 0.7203 &0.4726 & 0.6274 \tabularnewline
\hline%

\multirow{2}{12mm}{\centering WebKb}&80&0.8309 &0.3360 & 0.5931 \tabularnewline
 \cline{2-5}%
 & 90& 0.8583 &0.5736 & 0.6693 \tabularnewline
\hline%
\end{tabular}
\caption{Variation of accuracy with different training set sizes.}
\label{tab:sampling}
\end{figure}

\subsection{Importance of various features}

Figure~\ref{tab:CC} 
dissects the various features involved in
\OurAlgo, fixing at 90\% sampling rate henceforth.  To understand the
role of different features (local link structure, global community
structure), we build our SVM model eliminating either LL or CC and
calculate the ranking accuracy (MAP).  The results show that the
absence of each of the signals (LL and CC) significantly deteriorates
the performance.  However, closer scrutiny showed the interesting
property that the deterioration occurs in different zones, that is, it
is almost always true that the node whose MAP gets affected by
elimination of LL does not face such problem when CC is removed.

\begin{figure}
\centering \begin{tabular}{|p{1.4cm}|c|c|c|}
\hline%
 Dataset & \OurAlgo\  & $SVM(LL,AA)$ & $SVM(LL,CC)$ \\ \hline
{\centering Netflix}
 & 0.6017 & 0.5761 &0.5478\tabularnewline
\hline%
{\centering Movielens}
 & 0.8740  & 0.8436 & 0.8483 \tabularnewline
\hline%
{\centering Citeseer}
 & 0.7719 & 0.7467&0.7677 \tabularnewline
\hline%
{\centering Cora}
 & 0.7203 & 0.7044&0.7139 \tabularnewline
\hline%
{\centering WebKb}
 & 0.8583 & 0.8106&0.8470 \tabularnewline
\hline%
\end{tabular}
\caption{Feature ablation study.}  \label{tab:CC}
\end{figure}

\subsection{Effect of quadratic terms}

The results presented so far used only linear features.
Figure~\ref{tab:QT} shows the results obtained with quadratic features
(Section~\ref{sec:Discrim:Definition}), compared to linear features
(90\% sample).  The data sets are arranged in decreasing order of
size.  I.e., Netflix has the largest number of nodes and WebKb the
smallest.  It is seen that for larger data set the inclusion of
quadratic terms boosts accuracy.  The enhancement of performance
indicates the role played by possible interaction of the features.  In
other words, we can conclude that not only local link structure,
attributes and community level signals have important roles here, but
also they interact or affect each other which also play a crucial role
in link prediction.  The interaction can be exploited better in a more
complex model if there is more training data.

\begin{figure}
\centering \begin{tabular}{|p{1.4cm}|c|c|c|}
\hline%
 Dataset & \OurAlgo\  & LL+AA+CC & Improvement Factor\\ \hline
{\centering Netflix}
 & 0.6017 & 0.5755 & 1.045 \tabularnewline
\hline%
{\centering Movielens}
& 0.8740  & 0.8565 & 1.020 \tabularnewline
\hline%

{\centering Citeseer}
 & 0.7719 & 0.7677 & 1.006\tabularnewline
\hline%

{\centering Cora}
 & 0.7203 & 0.7220 & 0.997 \tabularnewline
\hline%

{\centering WebKb}
 & 0.8583 & 0.8524 & 1.006 \tabularnewline
\hline%
\end{tabular}
\caption{Effect of including quadratic features.}
\label{tab:QT}
\end{figure}

\subsection{ Workload Distribution}
\label{workLoad}

\begin{figure}
\begin{adjustwidth}{-0.5cm}{}
    \centering
   \includegraphics[width=.9\hsize]{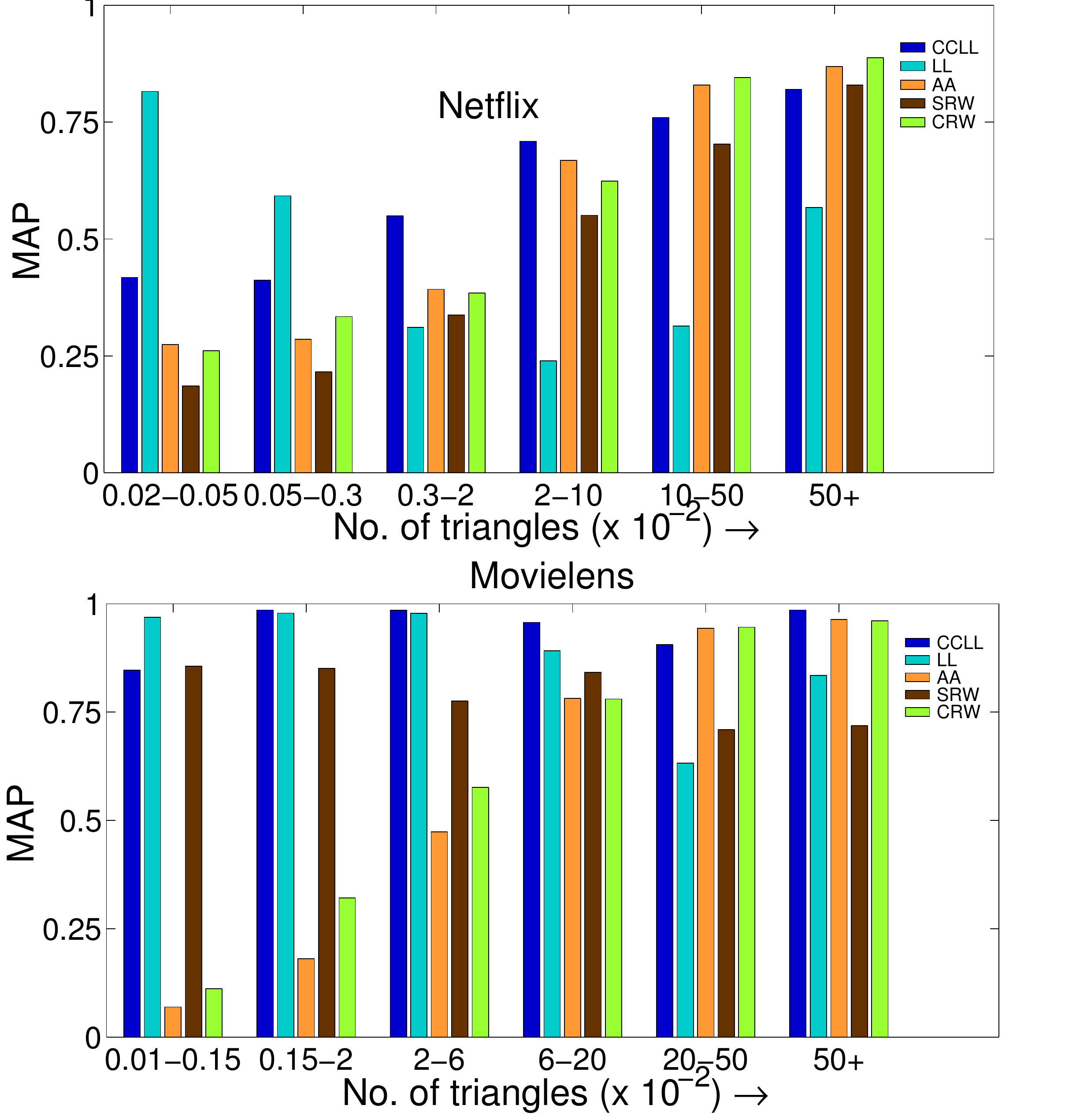}
   \caption{Workload distribution based on number of triangles.}
    \label{fig:wl_triangle}
\end{adjustwidth}
    \end{figure}

\begin{figure}
\begin{adjustwidth}{-0.5cm}{}
    \centering
  \includegraphics[width=.9\hsize]{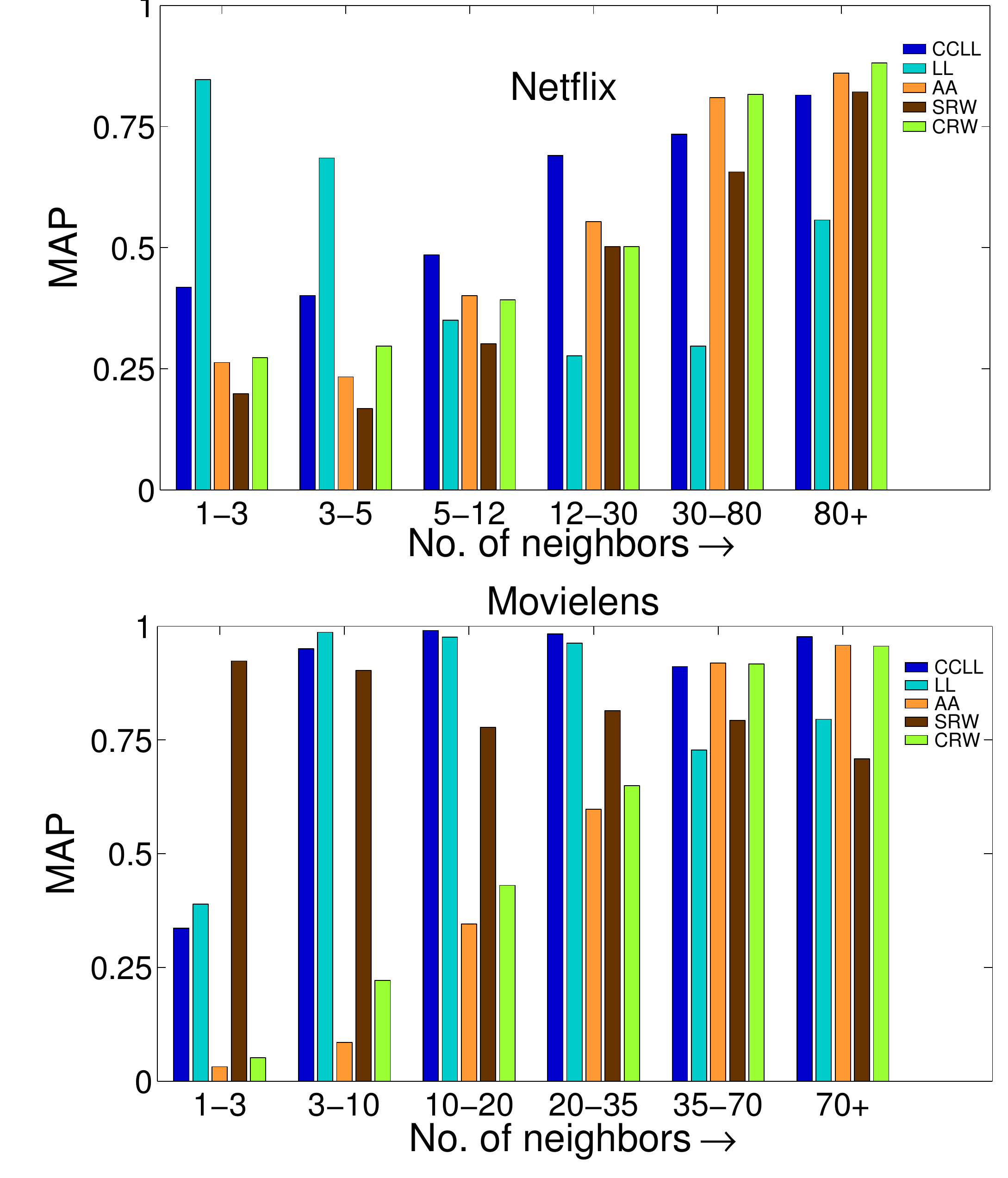}
   \caption{Workload distribution based on number of neighbors.}
    \label{fig:wl_neighbors}
\end{adjustwidth}
    \end{figure}

In this section, we present some comparative analysis between
\OurAlgo\ and other four best benchmark algorithms on two
representative datasets: Netflix and Movielens (Figures ~\ref{fig:wl_triangle} and ~\ref{fig:wl_neighbors}).   The choice is
motivated by the fact that Movielens gives best performance and
Netflix gives worst performance (with \OurAlgo) among all the five
datasets.  Netflix has few features while Movielens is feature-rich.
Query nodes are bucketed based on 
\begin{itemize}
\item the number of neighbors they have (changes from sparse to
  dense), and
\item the number of triangles formed.
\end{itemize}
Each bucket holds roughly one-sixth of the nodes.

The workload distribution highlights the nature of each algorithm.
The behavior of the algorithms is similar for the two workload
distributions.  It is seen that AA and CRW which solely depend on link
structure improve as the graph becomes more dense (number of neighbor
increases) or become more social (number of triangle increases).  The
two feature based algorithms, LL and SRW, perform well in the sparse
zone, and the improvement in the dense (more social) zone is observed
but not as significant as the two link-based algorithms.  Clearly, in
these two zones (sparse and dense), two different classes of algorithm
work well.


\OurAlgo\ performs well in all the zones by appropriately learning
signals in each zone.  However, it even improves upon its two
constituents in each and every zone.  From
Figure~\ref{fig:wl_neighbors} we observe that \OurAlgo\ performs best
(substantially better than LL and AA), at intermediate density.  There
are two reasons behind it.  Even though the graph is sparse, in these
regions, $|G(q)|$ and $|B(q)|$ are not far apart, which helps
\OurAlgo\ to train better.  Second, nodes in these zones are member of
community coclustering structures with informative block densities and
surprise feature.  Factoring in the community signal helps to
positively interpret the surprise.

%
%
%

\subsection{Variation of performance across different data sets}

\begin{figure*}
\begin{adjustwidth}{-0cm}{}
\includegraphics[width=\textwidth]{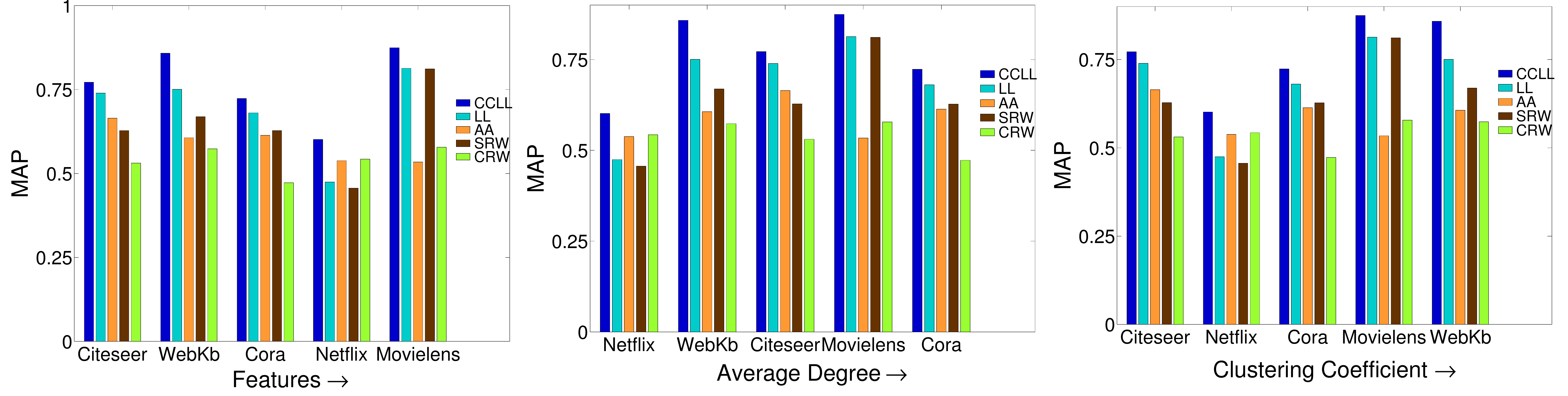}
    \caption{Variation of MAP over various datasets w.r.t.\ feature
      richness, average degree and clustering coefficient.
      (The same data is presented in three different orders.)}
    \label{fig:global_all}
    \end{adjustwidth}
\end{figure*}

From Figure~\ref{tab:results} we observe the wide variation of MAP
over various datasets, across all algorithms.  The
variation may be due to various graph properties like density,
clustering coefficient or due to the (un)structureness of the
constituent features.  We organize the results of MAP with respect to
the above mentioned three parameters and present the same through
Figure~\ref{fig:global_all}.  Following the 
first sub-figure, we see
Movielens is having the most well-structured feature space.  It
consists of genre and it is usually observed that people usually like
similar movies.  We find that it has the highest MAP.  
Feature richness is not a perfect predictor of high MAP.
Netflix has rich features but perform relatively poorly.  
Like MovieLens, Netflix is also a movie-movie
network.  But its features are not as informative, because they
are derived from many other factors (like year of the
movie, actors, director etc.) apart from genre, and the
distillation process seems noisy.

The other predictor of MAP is the average density (subfigure (b)), higher density means more signal to the classifier. 
Hence, MAP normally increases with increasing density,  
that explains Netflix which has very low average degree perform poorly. The third factor which enhances accuracy is local 
connectedness which is captured through clustering coefficient and is plotted in subfigure (c).   The three paper
repositories, WebKb, Cora and Citeseer, have very unstructured
feature space (words) as the features are polluted by polysemy, synonymy,
etc.  However, the high clustering coefficient of WebKb balances this
disadvantage and is a key reason for its good performance.  For the
other algorithms LL and SRW, the same ranking is maintained, although
the performance gap between Movielens (first) and WebKb (second)
increases.  This is in line with the same observation mentioned before
that LL and SRW cannot exploit link structure well. The ranking in CRW
and AA are different and more in line with the ranking of the dataset
with respect to density and clustering coefficient.

%% file: End.tex
\section{Conclusion}
\label{sec:End}

We described a new two-level learning algorithm for link prediction.
At the lower level, we learn a local similarity model across edges.
At the upper level, we combine this with co-clustering signals using a
SVM.  On diverse standard public data sets, the resulting link
predictor outperforms recent LP algorithms.  Another contribution of
this paper is to systematically understand the areas of inefficiency
of recent LP algorithms and consequently establish the importance of
this two-level learning scheme, and the key features we use.  We show
that our algorithm consistently outperforms four strong baselines when
link information is neither too sparse nor too dense.  In practice, a
large amount of requests for link recommendation will actually come
from this zone, hence the significance of the result can be even more
than what is stated in the paper.  In future work, it will be of
interest to combine the signals we exploit with supervised
personalized PageRank \cite{BackstromL2011SRW}.  Another possibility
is to replace item-wise or pairwise losses with list-wise losses
suitably aggregated over query samples.


%% file: Ack.tex
\vspace{0.1cm}
\label{sec:Ack}
\textbf{Acknowldegement: } {This work was
supported by Google India under the Google India PhD Fellowship Award.}